\documentclass[journal]{IEEEtran}

\usepackage{cite}
\usepackage{graphicx}
\graphicspath{{./}{photo/}{generated/phase1_figure3/}}
\DeclareGraphicsExtensions{.pdf,.png,.jpg,.jpeg}
\usepackage{amsmath,amssymb}
\usepackage{booktabs}
\usepackage{tabularx}
\usepackage{array}
\usepackage{multirow}
\usepackage{threeparttable}
\usepackage{url}
\newcommand{\figlabel}[1]{\textcircled{\scriptsize #1}}

\begin{document}

\title{KRVF: A Source-Aware Semantic Voxel World Representation for Edge Mobile Manipulation}

\author{Runfeng~Ling%
\thanks{Runfeng Ling is with The University of Manchester, Manchester, U.K.}}

\markboth{KRVF Technical Report}{Ling: KRVF}

\maketitle

\begin{abstract}
Mobile manipulators need world models that are current, queryable, semantically meaningful, and usable under edge-compute constraints. This technical report presents KRVF, a source-aware semantic voxel world representation for edge mobile manipulation. Unlike reconstruction-centric mapping pipelines that primarily optimize global geometric fidelity, KRVF represents local world state as task-oriented voxels that encode occupancy, color, semantic evidence, temporal freshness, and evidence source. The representation separates measured occupancy from semantic-prior hypotheses, enabling depth-failure-aware object reasoning without silently corrupting persistent geometry. KRVF also closes a feedback loop between mapping and sensing by rendering map-prior depth for repair, and exposes task-level query operators for semantic objects and grasp candidates. The report formalizes the KRVF representation and documents a ROS~2 implementation that turns online RGB-D observations into a task-facing robot memory.
\end{abstract}


\IEEEpeerreviewmaketitle

\section{Introduction}

\IEEEPARstart{R}{obotic} mapping is often framed as a reconstruction problem: given sensor observations and robot poses, estimate the geometry of the surrounding world. This framing is effective for navigation, localization, and offline scene reconstruction, but it is not always the right objective for mobile manipulation. A mobile manipulator operating on edge hardware does not merely need a complete map. It needs a representation that answers task-level questions with low latency: where is the object, how reliable is the evidence, is the observation still fresh, and what action can be attempted next?

This distinction becomes especially important in tabletop and near-field manipulation scenarios. RGB-D sensors may fail on transparent, reflective, high-gloss, dark, or thin objects. Dynamic objects may leave stale geometric remnants after being moved. A semantically correct detector may identify an object, while the underlying depth map provides incomplete or misleading geometry. Conversely, a geometrically valid point cloud may remain difficult for downstream manipulation modules to query in terms of semantic object identity or grasp affordances.

KRVF addresses this gap by treating the map as a source-aware, queryable robot world state. It fuses RGB-D observations into a sparse semantic voxel field, explicitly distinguishes measured geometry from semantic hypotheses, maintains freshness and confidence cues, and exposes object and grasp queries as first-class outputs. KRVF is not intended to replace SLAM backends; it assumes an external pose source such as TF, odometry, or a SLAM/localization system. Its role is to convert online perception into a world representation that manipulation and behavior modules can query directly.

The central idea is that a robot should track why it believes a region is occupied, whether that belief is still current, and whether the resulting structure can support a task. In KRVF, observed geometry, semantic labels, hypotheses, repaired depth, uncertainty, and grasp affordances are represented as different aspects of the same task-oriented voxel world.

The contributions of this report are fivefold:
\begin{enumerate}
\item KRVF is formulated as a task-oriented semantic voxel world representation for edge mobile manipulation, designed around action latency, semantic queryability, and graspability rather than global reconstruction fidelity.
\item Occupancy voxels are extended into source-aware task voxels that encode occupancy, color, semantic evidence, observation confidence, temporal freshness, and evidence type.
\item Measured occupancy is separated from semantic-prior hypotheses, allowing depth-failure-aware reasoning while preserving the integrity of persistent measured geometry.
\item A map-prior and depth-failure handling pathway is described, in which the current voxel map can be rendered back into the camera view as a depth prior for repair, and semantic-prior hypothesis updates are interpreted as a semantically gated extension to inverse sensor modeling.
\item The world model is exposed through semantic object and grasp-candidate query operators, allowing behavior-tree and manipulation pipelines to consume KRVF as a task-facing memory through a decoupled ROS interface.
\end{enumerate}

\section{Related Work and Positioning}

KRVF is positioned between occupancy mapping, dense reconstruction, RGB-D SLAM, semantic mapping, and manipulation-oriented perception. It is not intended to replace these systems. Its role is to provide a source-aware representation layer that turns online RGB-D observations into a queryable state for mobile manipulation.

\textbf{Occupancy mapping.}
OctoMap-style occupancy mapping provides a compact probabilistic model of free, occupied, and unknown space~\cite{hornung2013octomap}. This is useful for navigation and collision reasoning, but the map is usually consumed as geometry. KRVF keeps the log-odds view of occupancy, while extending each voxel with semantic evidence, freshness, source type, and task-level query access. The main difference is not the use of voxels, but the treatment of voxel state as evidence-bearing robot memory.

\textbf{Dense fusion.}
TSDF and Voxblox-like dense fusion systems are well suited to building locally consistent surfaces from RGB-D or depth observations~\cite{newcombe2011kinectfusion,oleynikova2017voxblox}. They are strong choices when the objective is surface quality, signed-distance queries, or dense geometric reconstruction. KRVF targets a different operating point: lower-latency local state, explicit source labels, disposable hypotheses, and direct object/grasp queries. It gives up some reconstruction ambition in exchange for a representation that is easier for an edge manipulation stack to consume at action time.

\textbf{RGB-D SLAM.}
RTAB-Map and related RGB-D SLAM systems estimate camera motion, build maps, and support loop closure or localization~\cite{labbe2019rtabmap}. KRVF assumes this kind of pose source may exist upstream. It does not solve global localization or loop closure. Instead, it consumes synchronized RGB-D data and TF/pose transforms, then maintains a local semantic voxel world state for manipulation. This separation is deliberate: SLAM estimates where the robot is, while KRVF organizes what the robot can currently act on.

\textbf{Semantic mapping.}
Semantic mapping systems attach class labels, object instances, or scene-level semantics to geometric maps~\cite{mccormac2017semanticfusion,rosinol2020kimera}. KRVF follows this direction, but emphasizes evidence source and task queries. A detector output is not treated as geometry by itself; it must be grounded into voxel evidence or stored as an explicitly marked hypothesis when depth is missing. This allows semantic information to support action without silently overwriting measured occupancy.

\textbf{Grasp-from-detection pipelines.}
Many practical manipulation systems map a 2D detection or segmented depth crop directly to a grasp pose~\cite{redmon2018yolov3,mahler2017dexnet}. Such pipelines can be effective, but they often keep perception as a transient frame-level computation. KRVF instead stores detections as part of a temporally maintained voxel state, allowing object queries, stability estimates, freshness checks, and grasp candidates to be generated from the current world model rather than from one image alone.

\textbf{Neural and Gaussian scene representations.}
Neural fields and 3D Gaussian scene representations offer powerful tools for view synthesis and high-fidelity scene modeling~\cite{mildenhall2021nerf,kerbl20233dgs}. They are less directly aligned with the constraints targeted here: online operation, explicit evidence source, low-latency ROS queries, and simple integration with behavior trees and manipulation pipelines. KRVF is therefore closer to a robot systems representation than a photorealistic scene model.

Across these lines of work, the central positioning is the same: KRVF is a representation layer for acting under imperfect sensing. It can use SLAM, RGB-D sensing, semantic detectors, and depth repair as inputs, while presenting manipulation modules with object and grasp queries instead of raw reconstruction outputs. This also gives KRVF a practical integration boundary: robot platforms that provide synchronized RGB-D observations, camera calibration, and a pose transform can use the same mapping core, while task execution remains outside the mapper.

\begin{table}[!t]
\centering
\caption{Qualitative positioning of KRVF against common robot perception representations. KRVF exposes source awareness, hypothesis separation, temporal freshness, object queries, and grasp queries as first-class representation features. The comparison clarifies interface-level capabilities rather than quantitative performance.}
\label{tab:positioning}
\scriptsize
\setlength{\tabcolsep}{2.2pt}
\renewcommand{\arraystretch}{1.12}
\begin{tabularx}{\columnwidth}{>{\raggedright\arraybackslash}X c c c c c}
\toprule
\textbf{Representation} & \textbf{Src.} & \textbf{Hyp.} & \textbf{Temp.} & \textbf{Obj.} & \textbf{Grasp} \\
 & \textbf{aware} & \textbf{sep.} & \textbf{fresh} & \textbf{query} & \textbf{query} \\
\midrule
PointCloud2 & No & No & Limited & No & No \\
OctoMap-style occupancy & Limited & No & Limited & No & No \\
RTAB-Map cloud & Limited & No & Limited & No & No \\
Direct YOLO-depth pipeline & No & No & Frame-level & Partial & Partial \\
KRVF & Yes & Yes & Yes & Yes & Yes \\
\bottomrule
\end{tabularx}
\end{table}

\section{KRVF World Representation}

KRVF represents the robot's local operating environment as a sparse semantic voxel field. Each voxel stores geometric, visual, semantic, temporal, and evidential state. The representation is designed for online use on edge hardware, where the objective is not to maintain the most complete global reconstruction, but to maintain a fresh and queryable local world state suitable for manipulation.

At time \(t\), the KRVF state can be viewed as a union of three interacting layers:
\begin{equation}
\mathcal{K}_t =
\mathcal{K}^{obs}_t
\cup
\mathcal{K}^{hyp}_t
\cup
\mathcal{K}^{free}_t ,
\label{eq:krvf_layers}
\end{equation}
where \(\mathcal{K}^{obs}_t\) is the observed voxel layer maintained from RGB-D evidence, \(\mathcal{K}^{hyp}_t\) is the semantic hypothesis layer used for bounded task hypotheses under missing or failed depth, and \(\mathcal{K}^{free}_t\) stores free-space evidence derived from conservative ray carving. This decomposition is a representational choice: it allows the robot to distinguish measured occupancy from semantic hypotheses and free-space evidence, rather than collapsing all evidence into one undifferentiated map.

For runtime efficiency, KRVF is also exposed through active and persistent views:
\begin{equation}
\mathcal{M}_t =
\mathcal{M}^{active}_t
\cup
\mathcal{M}^{persistent}_t
\cup
\mathcal{M}^{hyp}_t .
\label{eq:runtime_views}
\end{equation}
The active map \(\mathcal{M}^{active}_t\) is a high-frequency local window around the robot or camera. It supports immediate feedback for manipulation and visualization. The persistent map \(\mathcal{M}^{persistent}_t\) accumulates longer-lived observed geometry and can be published at a lower frequency. The hypothesis map \(\mathcal{M}^{hyp}_t\) provides explicit, disposable semantic-prior occupancy when physical depth is unreliable.

This design reflects a task-level assumption: not every downstream consumer needs the full voxel map at sensor frame rate. A mobile manipulator instead needs a high-frequency local state for action, a lower-frequency persistent memory for context, and an explicitly marked hypothesis channel for uncertain but task-relevant evidence.

\subsection{Pipeline Overview}

The KRVF pipeline takes synchronized RGB-D images, camera calibration, and a pose transform into an output frame. Valid depth samples are back-projected into 3D, transformed into the world frame, and inserted into a chunked sparse voxel map. Occupancy, color, confidence, and temporal information are fused per voxel. Free-space evidence is integrated through conservative ray carving, while dynamic clearing and temporal cleanup reduce stale or unsupported structures.

Semantic information enters through bridge nodes rather than being hard-coded into the mapper. A 2D detector such as YOLO provides class evidence, but KRVF grounds this evidence into existing 3D occupied voxels through depth sampling and frame transformation. When depth is missing or unreliable, semantic detections may create bounded hypothesis voxels in a separate overlay layer. Downstream consumers can query semantic objects and grasp candidates without directly interpreting raw point clouds.

\begin{figure*}[!t]
\centering
\includegraphics[width=0.95\textwidth]{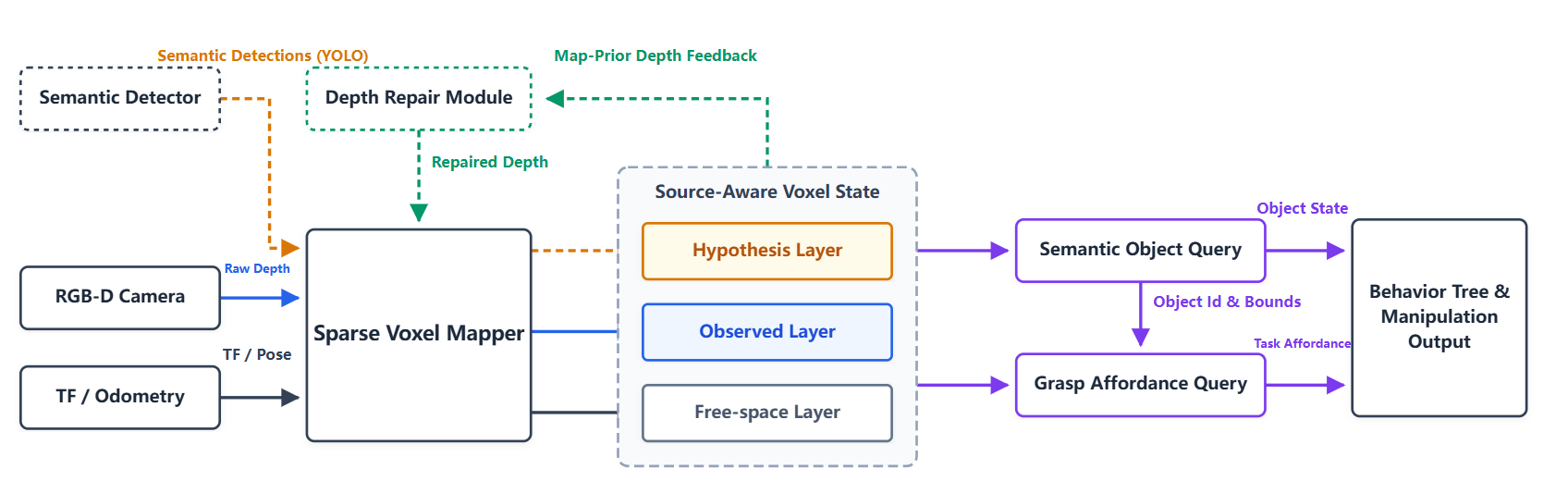}
\caption{KRVF pipeline overview. KRVF consumes synchronized RGB-D observations, camera calibration, and TF/pose input, and fuses them into a sparse source-aware voxel world state. The mapping core maintains observed occupancy, free-space evidence, and semantic hypothesis layers. Semantic bridge and depth-repair modules provide optional semantic updates and map-prior repair signals, while downstream behavior-tree and manipulation modules access the representation through object and grasp queries rather than raw point clouds.}
\label{fig:pipeline}
\end{figure*}

Operationally, KRVF converts RGB-D observations, pose, and semantic detections into source-aware semantic voxels, then into observed, hypothesis, and free-space layers. These layers are finally exposed as semantic objects and grasp affordance queries. The map is therefore consumed as a task-facing representation, instead of stopping at visualization.

\section{Source-Aware Voxel State}

Classical occupancy maps often reduce a voxel or cell to a geometric state:
\begin{equation}
v_i \in \{free, occupied, unknown\}.
\label{eq:classic_voxel}
\end{equation}
KRVF extends this into a source-aware task voxel:
\begin{equation}
v_i = (l_i, c_i, s_i, q_i, \tau_i, \eta_i),
\label{eq:krvf_voxel}
\end{equation}
where \(l_i\) is the occupancy log-odds, \(c_i\) is the visual appearance state, \(s_i\) is semantic evidence, \(q_i\) represents observation quality or source confidence, \(\tau_i\) is temporal freshness, and \(\eta_i\) denotes evidence type, such as observed, repaired, hypothesis, or uncertain.

This state definition is central to KRVF. The representation does not merely ask whether a voxel is occupied; it also records what evidence supports that belief and whether the belief remains actionable. A voxel derived from high-confidence raw depth should not be treated identically to a voxel produced by repaired depth or a semantic-prior hypothesis. Similarly, recently observed structure should not be treated identically to stale geometry left behind after a dynamic object has moved.

\subsection{Source-Weighted Evidence Fusion}

KRVF treats sensor updates as source-weighted evidence. A simplified positive occupancy update can be written as:
\begin{equation}
\Delta l_i = q_t \cdot L_{hit}(z_t, x_t, v_i),
\label{eq:source_weighted_update}
\end{equation}
where \(z_t\) is the current depth observation, \(x_t\) is the sensor pose, \(L_{hit}\) is the positive occupancy update for a hit voxel, and \(q_t \in [0,1]\) is the confidence associated with the observation source. This confidence may come from a depth confidence image, a repair confidence map, or a conservative default for raw depth.

This source weighting allows KRVF to integrate multiple observation modes without collapsing them into a single binary update. Raw depth, repaired depth, and semantic hypotheses can influence the world state with different confidence levels. In the implementation, confidence modulates log-odds integration and color fusion, while low-confidence observations are prevented from producing aggressive geometric updates.

\subsection{Appearance and Temporal State}

The appearance component \(c_i\) is also source-aware. KRVF maintains persistent and live color estimates: persistent color supports stable object identity, while live color reflects recent observations. A voxel color can be bootstrapped, frozen, blended, or recolored only after repeated evidence. This avoids allowing a single noisy frame to overwrite stable appearance while still permitting genuine scene changes.

The temporal component \(\tau_i\) is equally important. Each voxel tracks recent observation evidence, allowing the system to reason about freshness and apply cleanup to stale geometry. This temporal state supports dynamic manipulation environments where objects may be moved between observations and where outdated structure can become unsafe for action.

\section{Observed-Hypothesis Separation}

Depth sensors may fail exactly where manipulation needs reliable object state: transparent cups, reflective metal, glossy surfaces, thin objects, dark objects, or partially occluded items. A purely geometric occupancy update can treat missing or corrupted depth as free-space evidence, potentially underrepresenting or carving away the object that a semantic detector correctly identifies. Conversely, directly writing semantic detections into a persistent map risks polluting measured geometry with detector hallucinations.

KRVF addresses this through observed-hypothesis separation:
\begin{equation}
\mathcal{K}_t = \mathcal{K}^{obs}_t \cup \mathcal{K}^{hyp}_t,
\label{eq:obs_hyp}
\end{equation}
with the design principle
\begin{equation}
\mathcal{K}^{hyp}_t \not\rightarrow \mathcal{K}^{obs}_t
\label{eq:no_silent_promotion}
\end{equation}
unless later evidence explicitly promotes the hypothesis. The observed layer stores measured geometric evidence. The hypothesis layer stores semantic-prior occupancy that may be useful for task reasoning, visualization, or candidate generation, but is not silently fused into persistent geometry.

This distinction creates an epistemic boundary: the robot can represent what it measured separately from what it inferred. A semantic hypothesis can be visualized, queried, assigned a confidence, and used as a task candidate, while the persistent geometric map remains tied to measured or explicitly promoted evidence. This is particularly useful in depth-failure scenarios, where the observed map should preserve the fact that geometry is missing or unreliable, while the hypothesis layer can still mark a plausible object location for downstream reasoning.

\subsection{Non-Destructive Semantic Completion}

The hypothesis layer implements non-destructive semantic completion. Semantic priors are allowed to create actionable hypotheses, but they are not allowed to overwrite observed geometry by default. This prevents semantic detections from becoming permanent map facts without supporting evidence.

Hypotheses also have a temporal lifecycle:
\begin{equation}
h_i \xrightarrow[]{TTL} \emptyset .
\label{eq:hyp_expire}
\end{equation}
If not refreshed, a hypothesis expires. In future extensions, a hypothesis may be promoted to observed occupancy when confirmed by later valid depth or another trusted observation source:
\begin{equation}
h_i \xrightarrow[]{z_t\ valid} o_i .
\label{eq:hyp_promote}
\end{equation}
The first version of KRVF primarily implements explicit hypothesis overlays and expiration. The promotion rule is included as part of the broader representation design rather than claimed as a fully evaluated mechanism in this report.

\section{Map-Prior and Depth-Failure Handling}

KRVF includes two complementary mechanisms for handling unreliable depth: semantic-prior hypotheses and map-prior depth feedback. Semantic-prior hypotheses preserve explicitly marked task candidates when physical depth is missing or unreliable. Map-prior feedback allows previously observed geometry to support future depth repair. Together, these mechanisms allow the system to reason about depth failure without treating missing depth as either reliable free space or unquestioned occupied geometry.

\subsection{Semantic-Gated Inverse Sensor Interpretation}

Classical occupancy mapping can be written as
\begin{equation}
l_{i,t} = l_{i,t-1} + L_{phys}(z_t \mid x_t, v_i),
\label{eq:classic_inverse_sensor}
\end{equation}
where the update is driven primarily by the physical depth observation. KRVF introduces a semantic-prior hypothesis layer that can be interpreted as a semantically gated extension to this inverse sensor model:
\begin{equation}
\Delta l_i =
\alpha_i L_{phys}(z_t, x_t, v_i)
+
(1-\alpha_i)L_{sem}(S_t, x_t, v_i),
\label{eq:semantic_gated_inverse_sensor}
\end{equation}
where \(S_t\) is semantic evidence, \(L_{sem}\) is a semantic-prior update, and \(\alpha_i\) controls the relative trust assigned to physical and semantic evidence. When physical depth is trusted, the update is dominated by \(L_{phys}\). When physical depth is missing or unreliable but semantic evidence indicates a task-relevant object, the semantic term can create a bounded hypothesis rather than allowing the object region to disappear from the task representation.

In the current implementation, this idea is realized conservatively as a separate semantic hypothesis overlay rather than as a full replacement of the physical ray model. KRVF does not silently hallucinate geometry into the persistent map. Semantic evidence creates bounded, explicitly marked occupancy hypotheses that downstream task modules may inspect or query.

\subsection{Map-Prior Depth Feedback}

KRVF can render the current voxel map back into the camera view as a depth prior. This forms a feedback loop:
\begin{equation}
\mathcal{M}_{t-1}
\rightarrow
\hat{z}^{prior}_t
\rightarrow
z^{repair}_t
\rightarrow
\mathcal{M}_t .
\label{eq:map_prior_loop}
\end{equation}
The map prior \(\hat{z}^{prior}_t\) provides a depth estimate from previously observed voxel geometry under the current camera pose. A depth repair module can combine this prior with the current corrupted depth image, local color-aware interpolation, local plane fitting, and temporal prior information. The repaired depth can then be integrated back into the KRVF map with its own confidence.

This mechanism reframes the map as an active component of perception: the world representation can help repair the sensor stream that updates it. In practical terms, this supports depth-failure demonstrations where the system can maintain an explicit distinction between raw missing depth, repaired depth, uncertain regions, and semantic hypotheses. It also provides a useful design principle for edge robots: historical world state can be used as a bounded prior for current perception, provided that the repaired observations remain source-labeled and confidence-weighted.

\section{Task-Level Object and Grasp Queries}

KRVF exposes the map through task-level query operators. Rather than requiring downstream modules to parse raw point clouds, KRVF clusters semantic voxels into objects and generates grasp affordance candidates. This query interface is part of the representation: it defines how the robot consumes the map for action.

Semantic object query can be written as
\begin{equation}
Q_{obj}(s, x) \rightarrow O_j,
\label{eq:object_query}
\end{equation}
where \(s\) is a semantic label or label name, \(x\) is an optional query origin or strategy input, and \(O_j\) is a semantic object. A KRVF object contains an object id, semantic label, semantic confidence, center, bounding box, mean color, voxel count, and stability score. The stability score summarizes spatial support and repeated semantic evidence, allowing downstream modules to prefer objects that are both geometrically and semantically supported.

Grasp query can be written as
\begin{equation}
Q_{grasp}(O_j)
\rightarrow
\{g^{top}, g^{waist}, g^{side}, g^{center}\}.
\label{eq:grasp_query}
\end{equation}
The current implementation derives top, waist, side, and center/fallback grasp candidates from the semantic voxel component geometry. Candidates are ranked by confidence derived from object stability, voxel count, and geometric heuristics. This is not intended as a general optimal grasp planner; it is a task-oriented affordance layer that converts semantic voxel objects into actionable grasp targets for downstream execution.

\subsection{Detection-to-Voxel Grounding}

KRVF does not treat 2D detector outputs as geometry. A semantic bridge samples depth points inside detection boxes, rejects unstable depth-edge samples, transforms the surviving points into the KRVF frame, and updates semantic labels only near existing occupied voxels. In other words, the detector does not create objects; it grounds labels onto existing voxel evidence.

The result is a grounded semantic map: object category evidence comes from visual detection, but object geometry and grasp position come from KRVF voxel clusters. This reduces the risk of directly executing manipulation from single-frame detector coordinates and keeps semantic evidence tied to 3D support.

\subsection{Query Interface for Robot Behavior}

KRVF provides services for object queries and grasp candidate queries, and an action-style interface for waiting on semantic objects. This makes the world model usable by behavior trees~\cite{colledanchise2018bt} or task planners. A behavior module can wait for an object label to appear, receive feedback about observed object count, and continue once a semantic object satisfying the query constraints is available.

This interface is part of the representation design. KRVF is a queryable state server for manipulation tasks, not just a map publisher. In this sense, the map becomes a robot memory with task-level access patterns, avoiding the need for every downstream component to reinterpret raw geometry.

\section{System Implementation}

KRVF is implemented as a set of ROS~2 packages, following the standard ROS communication model of decoupled nodes, topics, services, and actions~\cite{quigley2009ros}. The implementation separates the real-time voxel mapper, interface definitions, semantic bridges, and depth-failure utilities:
\begin{itemize}
\item \texttt{krvf\_mapping\_cpp}: high-performance C++ RGB-D voxel mapper.
\item \texttt{krvf\_mapping}: Python reference/prototype pipeline.
\item \texttt{krvf\_mapping\_interfaces}: messages, services, and actions.
\item \texttt{krvf\_semantic\_bridge}: semantic grounding, hypothesis bridge, grasp executor bridge, and behavior-tree query wrapper.
\item \texttt{krvf\_depth\_repair\_cpp}: depth failure injection, depth repair, uncertain surface publication, and hypothesis tests.
\end{itemize}

The C++ mapper is the primary runtime component. It uses a sparse chunked voxel structure and maintains observed, free-space, candidate, and hypothesis voxel stores. Its publishers are organized around the representation views described earlier: high-frequency active local clouds, lower-frequency persistent clouds, dirty chunk updates, semantic debug clouds, semantic object arrays, grasp markers, map statistics, and map-prior depth images. Subscriber-aware publishing and cache-dirty chunk rebuilds reduce unnecessary computation when outputs are not being consumed.

The semantic bridge package keeps detector-specific logic outside the mapper. It converts 2D detections into grounded semantic point updates, sends bounded hypothesis boxes for depth-failure cases, refreshes visible grasp candidates, and can forward KRVF grasp results to downstream manipulation topics. The depth repair package provides controlled failure injection, multi-source repair, confidence publication, and uncertain surface outputs for evaluating the representation under missing-depth conditions.

\subsection{Decoupled Robot Integration}

KRVF is designed around a small integration contract rather than a specific robot platform. The mapping core requires synchronized RGB-D images, camera calibration, and a transform from the camera frame into the selected world or robot frame. It does not assume a particular arm, gripper, mobile base, detector, or motion planner.

Robot-specific logic is kept in bridge nodes. A semantic bridge adapts detector outputs into KRVF semantic updates. A grasp bridge converts KRVF grasp candidates into object poses or manipulation requests expected by the downstream robot stack. A behavior-tree wrapper exposes waiting and query behavior through actions and services. As a result, the mapper can remain stable while the detector, behavior tree, MoveIt pipeline, or hardware driver changes.

This decoupling is a practical part of the representation. KRVF is intended to be portable across RGB-D robots: replacing the robot should mainly require remapping camera topics, TF frames, semantic labels, and downstream grasp execution topics, not rewriting the voxel world model.

Key outputs include \texttt{/krvf\_cpp/active\_cloud}, \texttt{/krvf\_cpp/cloud}, \texttt{/krvf\_cpp/dirty\_cloud}, \texttt{/krvf\_cpp/semantic\_cloud\_debug}, \texttt{/krvf\_cpp/semantic\_objects}, \texttt{/krvf\_cpp/grasp\_candidate\_markers}, \texttt{/krvf\_cpp/map\_prior\_depth}, and \texttt{/krvf\_cpp/map\_stats}. Key services and actions include \texttt{/krvf\_cpp/update\_semantic\_points}, \texttt{/krvf\_cpp/update\_semantic\_region}, \texttt{/krvf\_cpp/update\_hypothesis\_box}, \texttt{/krvf\_cpp/query\_object}, \texttt{/krvf\_cpp/query\_grasp\_candidates}, and \texttt{/krvf\_cpp/wait\_for\_object}.

\section{Demonstrations and Qualitative Results}

The first KRVF demonstration set focuses on qualitative evaluation of the representation and task pipeline. The goal of this report is technical documentation and system formalization rather than a complete benchmark study. Each demonstration is intended to illustrate a specific property of the proposed world representation.

All demonstrations in this first report are conducted in simulation or replay-style ROS~2 experiments. No physical robot experiments are claimed in this version. This scope is intentional: the report archives the representation, interfaces, and qualitative behavior of the system before hardware deployment. Real-robot validation is left for future work.

\begin{figure*}[!t]
\centering
\includegraphics[width=0.92\textwidth]{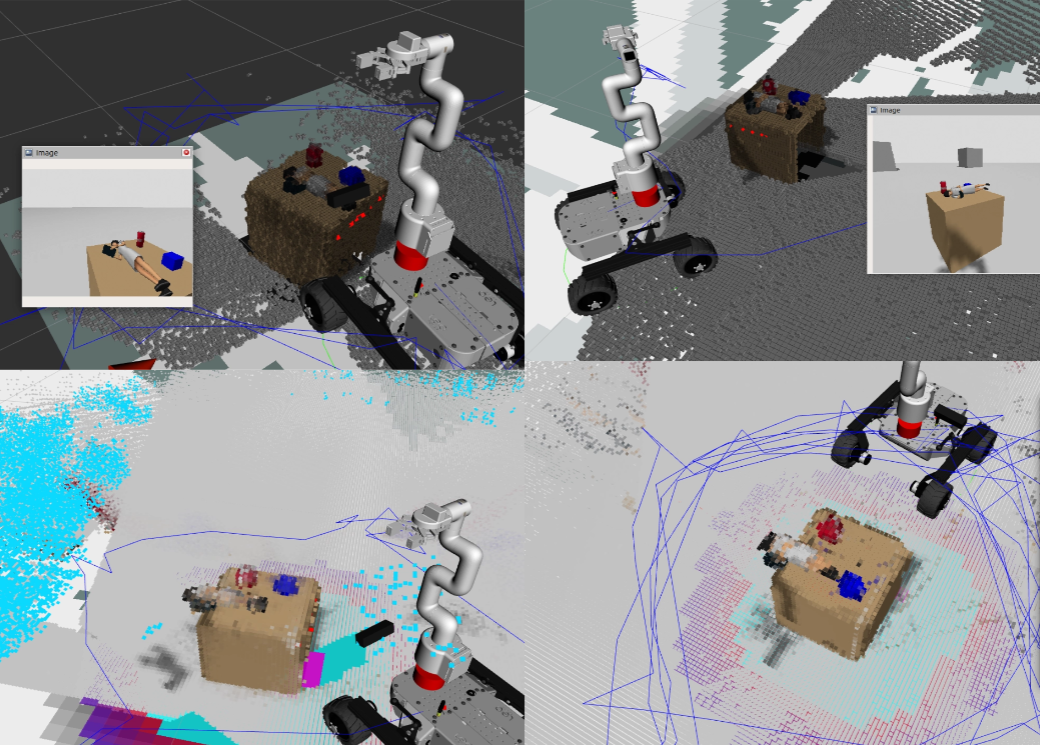}
\caption{Qualitative KRVF runtime views. Four RViz views of the same KRVF map representation are shown. The two upper panels show the block-style world rendered from different viewpoints using MarkerArray cube markers, emphasizing the discrete voxel-block structure used for visual inspection. The two lower panels show the corresponding voxel point-cloud representation from two viewpoints, emphasizing the dense spatial distribution of occupied voxels. Together, the panels show that KRVF can expose the same voxel state both as block-level visualization and as point-cloud geometry.}
\label{fig:runtime_views}
\end{figure*}

\subsection{Real-Time RGB-D Voxel Mapping}

The mapper produces a high-frequency active voxel cloud that follows the robot and a persistent map that accumulates observed geometry over time. This demonstrates the active/persistent split: the system can provide immediate local feedback while retaining longer-lived observed structure.

\subsection{Semantic Grounding}

YOLO detections are projected into the KRVF frame and grounded onto occupied voxel clusters. Semantic object markers and semantic debug clouds show that category labels are attached to 3D evidence rather than raw detector outputs. This supports detection-to-voxel grounding and the separation between visual category evidence and geometric support.

\subsection{Depth Failure and Hypothesis Overlay}

In a depth-failure scenario, the observed map shows missing geometry in the corrupted region, while the semantic hypothesis layer creates a bounded overlay for the detected object. The persistent observed map remains unpolluted. This illustrates observed-hypothesis separation and non-destructive semantic completion.

\begin{figure*}[!t]
\centering
\includegraphics[width=0.92\textwidth]{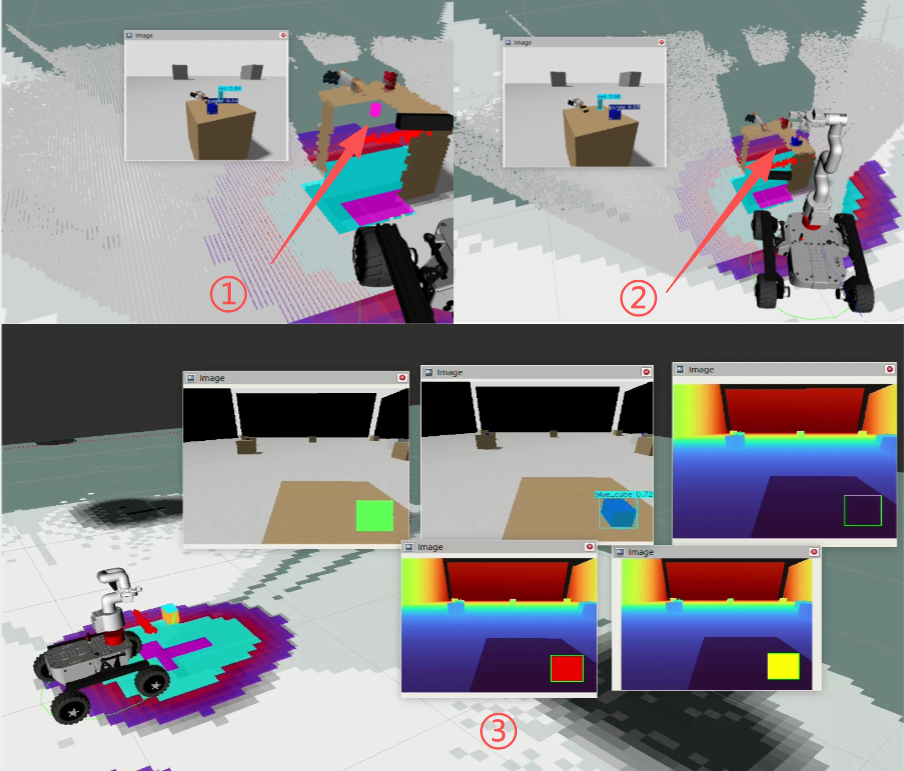}
\caption{Depth-failure handling with semantic completion. Three-panel visualization of KRVF behavior around a depth-failure region. \figlabel{1} The upper-left panel shows the target after it is moved into the depth-failure space; KRVF creates a block-level semantic hypothesis in real time even though the physical depth observation is unreliable. \figlabel{2} The upper-right panel shows the target next to the depth-failure space before entering it, where normal observed geometry remains available and no semantic completion is required. \figlabel{3} The lower panel shows a controlled failure case in which the depth values inside the YOLO detection box are manually removed; KRVF creates an explicitly marked queryable grasp target from the semantic evidence and the voxel-world prior. The comparison illustrates that KRVF can create an explicit task-level completion for the failed region while keeping this completion distinguishable from ordinary observed geometry.}
\label{fig:depth_failure}
\end{figure*}

\subsection{Grasp Candidate Query}

For semantic voxel objects, KRVF generates grasp candidate markers such as top, waist, side, and fallback grasps. This demonstrates that the map can be queried as an action representation rather than only visualized as geometry. The grasp markers are derived from voxel object geometry and stability rather than from the 2D detection alone.

\subsection{KRVF-to-Manipulation Bridge}

The grasp executor bridge queries KRVF for the best visible semantic object and publishes an object pose for downstream manipulation execution. This completes the pipeline from RGB-D observation to semantic voxel object to grasp target, showing how KRVF can operate as a task-facing world state server.

\section{Discussion}

KRVF targets a different regime from high-fidelity reconstruction systems. TSDF fusion, neural scene representations, and Gaussian scene models are designed primarily to optimize geometric or photometric quality. KRVF instead prioritizes low-latency, source-aware, queryable world state for edge mobile manipulation.

This does not make reconstruction-centric methods obsolete; rather, it defines a complementary representation layer. A robot may still use SLAM, localization, or dense reconstruction as supporting modules. KRVF focuses on the interface between online perception and task execution, where stale geometry, missing depth, semantic uncertainty, and query latency directly affect manipulation behavior.

The observed-hypothesis separation is particularly important. Semantic priors are useful, but they should not silently overwrite measured geometry. By keeping hypotheses explicit and disposable, KRVF allows semantic evidence to support action while preserving the integrity of the observed map. This design also makes failures more inspectable: a missing-depth region, a repaired-depth estimate, an uncertain surface, and a semantic hypothesis can be represented differently rather than collapsed into a single occupancy value.

\section{Limitations and Future Work}

KRVF is not a full SLAM backend and depends on external pose information from TF, odometry, or localization. The semantic hypothesis layer is not a general physical reconstruction method for transparent or reflective objects; it is a bounded task hypothesis mechanism. Grasp candidate generation is heuristic and task-oriented rather than a complete grasp planning framework.

The current evaluation is limited to simulation and ROS~2 demonstration scenarios. It does not yet include real-robot manipulation trials, physical RGB-D sensor noise on hardware, calibration drift, gripper execution errors, or closed-loop grasp success measurements. These real-world factors are important for validating the full mobile manipulation claim and should be evaluated in future hardware experiments.

The current report emphasizes system formalization and qualitative demonstration. Future work should include quantitative evaluation of update latency, semantic query latency, grasp candidate quality, stale-object cleanup time, CPU usage on edge hardware, and success rates under depth-failure conditions. Additional work should also modularize the C++ implementation, add unit and integration tests, and evaluate KRVF against baselines such as raw PointCloud2, OctoMap-style occupancy mapping, RTAB-Map point clouds, and direct YOLO-depth grasping.

Future versions should also study promotion and demotion policies for hypotheses. The current system supports explicit hypothesis overlays and expiration; a more complete framework could promote hypotheses into observed state after trusted confirmation and demote unsupported hypotheses based on task-specific risk.

\section{Conclusion}

KRVF reframes robotic mapping as a source-aware, queryable world representation for manipulation. By separating observed geometry from semantic hypotheses, weighting evidence by source confidence, feeding map priors back into depth repair, and exposing object and grasp queries, KRVF provides a task-facing robot memory for edge mobile manipulation.

The key idea is not that every voxel must be geometrically perfect, but that every task-relevant belief should be represented with its source, freshness, confidence, and action interface. This makes KRVF a practical step toward robotic world representations that support acting under uncertainty rather than merely reconstructing the scene.

\appendices

\section{Dynamic Clearing}

KRVF includes dynamic clearing mechanisms to reduce stale-object risk. Free-space evidence does not immediately delete occupied voxels. Instead, occupied voxels accumulate missed observations, and their log-odds decay only after sufficient consecutive misses:
\begin{equation}
l_{i,t+1} = l_{i,t} - \beta \cdot m_i .
\label{eq:dynamic_clearing}
\end{equation}
The required number of misses can depend on local support:
\begin{equation}
m_i^{required} =
\begin{cases}
m_{stable}, & \text{near stable support} \\
m_{default}, & \text{otherwise}.
\end{cases}
\label{eq:miss_support}
\end{equation}
This allows isolated noise to disappear quickly while protecting supported object structure from accidental carving.

\section{Shadow-Shell Guard}

Lighting artifacts can produce dark duplicate surfaces near stable colored geometry. KRVF includes an appearance-aware shadow-shell guard that compares hue, saturation, and brightness between a new candidate voxel and nearby stable voxels. If a new observation appears to be a darker shell of an existing colored surface, it can be rejected rather than inserted as new geometry. This is an engineering robustness mechanism, not a core theoretical contribution. It supports the broader goal of preventing visually induced artifacts from becoming false geometry.

\section{Color Stabilization}

KRVF maintains persistent and live color state. Colors are bootstrapped from early observations, frozen after sufficient support, and recolored only after repeated evidence. Local median smoothing and color-edge rejection reduce visual noise. This stabilizes voxel appearance while preserving the ability to adapt to genuine scene changes.

\section{Candidate-to-Stable Promotion}

Candidate voxels provide a short-lived buffer for uncertain observations before they become stable map state. A candidate voxel may be promoted after sufficient repeated evidence:
\begin{equation}
v_i^{cand} \xrightarrow[]{N_{support}} v_i^{stable}.
\label{eq:candidate_promotion}
\end{equation}
This mechanism is intended to reduce edge jitter and transient false positives. In the first KRVF configuration, the candidate layer is optional; it is included here as part of the representation design and implementation archive.

\section{Explicit Uncertainty Surfaces}

The depth repair module can publish unresolved depth holes as uncertain surface points rather than silently treating them as free or occupied. This supports the epistemic framing of KRVF: missing depth is not the same as free space, and failed repair is not the same as measured geometry. This output is mainly useful for debugging, visualization, and future task policies that may reason explicitly about unresolved sensing failures.

\section{ROS Topics and Services}

The primary C++ mapper publishes active local clouds, persistent clouds, dirty chunk updates, semantic debug clouds, semantic object arrays, grasp candidate markers, map-prior depth images, and map statistics. Its main service and action interfaces update semantic evidence, update hypothesis boxes, query objects, query grasp candidates, and wait for semantic objects. These interfaces are part of the decoupling design: the mapper exposes world state and query operators, while detector-specific logic, behavior-tree logic, and robot execution remain outside the mapping core.

\end{document}